\documentclass[sigplan,screen]{acmart}
\AtBeginDocument{%
  \providecommand\BibTeX{{%
    \normalfont B\kern-0.5em{\scshape i\kern-0.25em b}\kern-0.8em\TeX}}}

\setcopyright{acmcopyright}
\copyrightyear{2023}
\acmYear{2023}
\acmDOI{XXXXXXX.XXXXXXX}

\acmConference[MM '23]{}{Oct 29 - Nov 03,
  2023}{Ottawa, CA}
\acmPrice{15.00}
\acmISBN{978-1-4503-XXXX-X/18/06}

\usepackage{graphicx}
\usepackage{subcaption}
\usepackage{lipsum} % for dummy text
\usepackage{wrapfig} % for wrapping the figure
\usepackage{booktabs} % for tabs spacing

\newcommand{\tit}[1]{\smallbreak\noindent\textbf{#1.}}

\begin{document}

\title{Let's ViCE! Mimicking Human Cognitive Behavior in Image Generation Evaluation}

% Human-inspired Image generated quality assessment by visual concepts 
%Towards Automatic Human-aligned Evaluation for Synthetic Images}
% Replicating Human Cognitive Process for Evaluating Generated Images via Visual Concepts
% Incorporating Visual Concepts for a Human-Like Evaluation Approach in Image Generation
% Human-Like Image Generation with Visual Concepts

\author{Federico Betti}
\email{federico.betti@unitn.it}
\affiliation{%
  \institution{University of Trento}
  \city{Trento}
  \country{Italy}
}

\author{Jacopo Staiano}
\email{jacopo.staiano@unitn.it}
\affiliation{%
  \institution{University of Trento}
  \city{Trento}
  \country{Italy}
  }

\author{Lorenzo Baraldi}
\email{lorenzo.baraldi@phd.unipi.it}
\affiliation{%
  \institution{University of Pisa}
  \city{Pisa}
  \country{Italy}}

\author{Lorenzo Baraldi}
\email{lorenzo.baraldi@unimore.it}
\affiliation{%
  \institution{University of Modena and Reggio Emilia}
  \city{Modena}
  \country{Italy}}

\author{Rita Cucchiara}
\email{rita.cucchiara@unimore.it}
\affiliation{%
  \institution{University of Modena and Reggio Emilia}
  \city{Modena}
  \country{Italy}}

\author{Nicu Sebe}
\email{nicu.sebe@unitn.it}
\affiliation{%
  \institution{University of Trento}
  \city{Trento}
  \country{Italy}
}

\renewcommand{\shortauthors}{Betti, et al.}

\begin{abstract}
Research in Image Generation has recently made significant progress, particularly boosted by the introduction of Vision-Language models which are able to produce high-quality visual content based on textual inputs.
Despite ongoing advancements in terms of generation quality and realism, no methodical frameworks have been defined yet to quantitatively measure the quality of the generated content and the adherence with the prompted requests: so far, only human-based evaluations have been adopted for quality satisfaction and for comparing different generative methods.
We introduce a novel automated method for \textit{Visual Concept Evaluation} (ViCE), i.e. to assess consistency between a generated/edited image and the corresponding prompt/instructions, with a process inspired by the human cognitive behaviour.
ViCE combines the strengths of Large Language Models (LLMs) and Visual Question Answering (VQA) into a unified pipeline, aiming to replicate the human cognitive process in quality assessment. This method outlines visual concepts, formulates image-specific verification questions, utilizes the Q\&A system to investigate the image, and scores the combined outcome.
Although this brave new hypothesis of mimicking humans in the image evaluation process is in its preliminary assessment stage, results are promising and open the door to a new form of automatic evaluation which could have significant impact as the image generation or the image target editing tasks become more and more sophisticated.
\end{abstract}





\received{20 February 2007}
\received[revised]{12 March 2009}
\received[accepted]{5 June 2009}

\maketitle

\section{Brave Idea Introduction}
Quantitatively assessing the results of image generation models is a complex task. The challenge is clear when we consider simpler generative models such as Variational Autoencoders (VAEs), and it escalates when we delve into unsupervised or self-supervised models like Generative Adversarial Networks (GANs) or Diffusion Models~\cite{sohl2015deep, dhariwal2021diffusion}. Often, the evaluation of proposed models is based on some measurements in the latent space or on specific features extracted from the generated images~\cite{hessel2021clipscore,banerjee_meteor_2005}; sometimes it is associated with checking the sharpness and diversity of results in comparison to a reference test dataset~\cite{heusel_gans_2018}. Thus, whether the model generates images from a single text prompt or modifies an existing image based on a textual input (a process commonly referred to as Image Target Editing), accurately gauging their effectiveness is an ongoing challenge.

In fact, thus far, the only universally agreed upon evaluation methodology is the ultimate human judgment.

In recent years, new models and commercial products have been introduced that offer unprecedented perceptual quality and realistic representation. Systems for prompt-based generation, such as those based on Diffusion Models~\cite{sohl2015deep,ho2020denoising, balaji2022ediffi}, and multimodal models that allows for partial modification of the input image~\cite{brooks2023instructpix2pix,zhang_hive_2023}, at first glance, seem to deliver satisfactory results. However, appearances can be misleading.

Given that the research community largely agrees that metrics are necessary for benchmarking these generative process, the key question here is: how can we evaluate the effectiveness of a generative process triggered by text or multimodal input without ground truth? This includes other underlying inquiries like: does the output image meet perceptual and semantic expectations? Does the output image meet the constraints of the textual prompt? Does the image accurately reflect changes requested during a generative editing process in a multimodal setting? 
Until now, no reference-less quantitative scoring framework has been proposed.

\begin{figure}
    \centering
    \includegraphics[width=.40\textwidth]{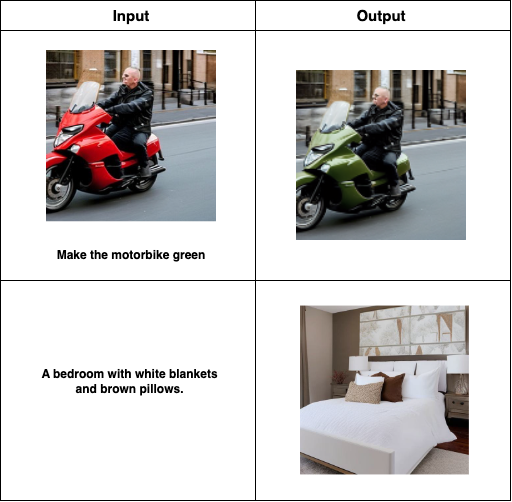}
    \caption{Different types of Image Generation tasks. \emph{Top}: in a multimodal Image Targeted Editing setup, given an input image paired with a textual instruction, the generative system is called to modify the former according to the latter. \emph{Bottom}: in a cross-modal image generation setup, the generative system is called to produce an image based on the textual description provide as the sole input.}
    \label{fig:generation_types}
\end{figure}

Let's consider a situation where we ask a generative system to partially modify the content of a given image: we could request to only change the color of the motorbike to green
(see Figure~\ref{fig:generation_types}). Here, a concrete challenge lies in determining whether the system can fulfill these kind of requests without introducing unintended alterations, and still effectively implement the desired modifications and maintain a high level of perceptual quality. In other words, \emph{how can we evaluate whether the system has precisely executed the requested changes, neither exceeding nor falling short, and has produced an aesthetically pleasing output?}

Right now, human input is crucial for this process. This involves either having humans annotate data beforehand or getting humans to assess the result after the generative process. Such human feedback can also be harnessed to continuously improve the models, e.g. through reinforcement learning methods~\cite{ouyang2022training}.

Despite the complex task of perfectly imitating human judgment, we can aim to emulate the strategy employed by humans to make judgments, mainly by asking and answering questions. This provides a viable approach to modelling human judgement within an AI system. This is the essence of what we term as human-aligned Visual Concept Evaluation (ViCE).

To sum up, our contributions include:
\begin{itemize}
    \item a novel interpretation method for images based on question answering, that reflects the human cognitive process;
    \item a universal evaluation protocol applicable to all image generation tasks, including Image Targeted Editing (ITE);
    \item an AI system, which leverages Large Language Models (LLMs) for dynamic question generation, which circumvents reliance on a static question pool;
    \item a semantic complement to perceptual quality metrics, contributing additional depth to evaluations, rather than attempting to replace existing metrics;
    \item enhanced alignment with human evaluations, bolstering the trustworthiness and authenticity of AI-generated assessments.
\end{itemize}
Embarking in this direction is indeed bold, as it seeks to bridge the cognitive gap between artificial intelligence systems and humans.

\section{Related Works}
\tit{Text-to-Image generation and editing} Over the past few years, many approaches have emerged in the realm of Generative AI, aiming to enhance the efficacy of image-generation tasks. Notably, advancements in the application of Generative Adversarial Network (GAN)~\cite{goodfellow2014generative,zhang2017stackgan,sauer2023stylegan,tao2023galip} and Diffusion Model~\cite{sohl2015deep,kingma2016improved,ho2020denoising,dhariwal2021diffusion,ramesh2022hierarchical, balaji2022ediffi} have significantly elevated the current state-of-the-art with regard to the Text-to-Image paradigm, which involves the generation of an image from a given textual description (or prompt). For example, in~\cite{rombach2022high} each step of the diffusion process is conditioned on the textual prompt input by the user, resulting in an output aimed at representing the starting textual concept. 

Given the significant advances in this field, more recent efforts have enlarged the scope of the Text-to-Image paradigm to encompass human-written instructions for image editing~\cite{brooks2023instructpix2pix,kawar2023imagic,zhang_hive_2023}. In this particular task, the objective is to manipulate the semantics of an image using a textual prompt, while simultaneously avoiding any undesired alterations to the image itself.

\tit{Metrics for Automatic evaluation of image generation and editing} 
Despite the significant efforts by the research community to enhance the qualitative outcomes of image editing and generation, only a limited number of techniques have been proposed to effectively evaluate the produced results of both methods.

As emphasized in~\cite{otani2023toward}, current automatic metrics exhibit limited performance in evaluating Text-to-Image generation when compared to human evaluations. Metrics such as Fréchet Inception Distance (FID)~\cite{heusel_gans_2018} and Inception Score (IS)~\cite{salimans2016improved} primarily focus on assessing image fidelity, disregarding the alignment between the generated image and the associated text. Conversely, CLIPScore~\cite{hessel2021clipscore} aims to measure the cosine similarity between the image and text tokens that are tokenized using CLIP image and text encoders. However, there are instances~\cite{gal2022stylegan,pmlr-v162-nichol22a} where generative models employ this metric to optimize image generation during training, leading to potential biases and unfair measurements at evaluation time.

To address this challenge, \cite{otani2023toward} propose a solution involving human evaluation as the primary method of evaluating Text-to-Image models. Further, a recently proposed automatic metric, LLMScore~\cite{lu2023llmscore}, despite combining global and local descriptions using a Large Language Model (LLM) into an object-centric visual description, presents some limitations. 

A significant drawback is that the generated captions often contain additional details that are not sourced from the image captioners, but instead fabricated by the LLM. Moreover, the final caption does not sufficiently incorporate the requirements and inputs from the original prompt, differing significantly from a human-like evaluation process. Eventually, LLMScore compares this description with the textual prompt used during the generation process and utilizes an LLM to compute the final score. Our proposed metric strives to overcome these issues and to more effectively replicate human visual reasoning -- by incorporating a pipeline that specifically evaluates the extent to which the generated image fits the textual requests.

Our work shares some commonalities with QuestEval~\cite{scialom_questeval_2021}, which implemented a similar strategy for text summarization tasks. In QuestEval, concepts crucial to the content were identified by means of question generation and question answering, and the summarized output was then evaluated based on the presence/absence of the same question/answer pairs. 

\section{Visual Concept Evaluation}
The process of Visual Concept Evaluation, as we define it, aims to replicate human behavior during the assessment of a generated image. When a human is asked to rate, on a scale of 1 to 10, how well an image generation task has been executed, his/her brain unconsciously starts considering the "visual concepts" they expect to see within the generated image. Visual concepts go beyond basic elements, such as shapes and colors, and include complex aspects such as specific objects and their contextual interaction within a scene.

These concepts are dictated by the initial text that forms the basis for the image generation in the case of traditional image generation tasks. However, for multimodal inputs, these concepts hinge on both the text and the input image. Additionally, evaluators utilize their implicit knowledge to infer other intuitive aspects.
For instance, if the prompt is "a cat on the stairs", the evaluator expects to see a cat, of which 1 to 4 legs might be visible, with the paws placed on the steps and a tail. All this information is easily deduced by a human brain and corresponds to the thought process a person goes through when assessing an image.

Hence, it is crucial to encapsulate not only the explicit instructions derived from the prompts, but also the implicit assumptions and expectations that humans naturally make. This brings forth the intricacy of the challenge - it's about recognizing and integrating these nuanced aspects of human cognition into the evaluation framework. Such broader understanding forms the foundation of the Visual Concept Evaluation process and is the key to aligning AI systems closer to human-like image assessment capabilities.
To represent the creation of the visual concepts, which we denote as $v_i$, we can use the following formulas.

Visual concepts are generated from the text $T$, and, in case of ITE task, the input image $I_{input}$. 

The visual concepts can be represented as:
\begin{equation}
\label{eqn:vo_from_text}
\begin{split}
V_{T} &= f(T) = {v_{1}, v_{2}, ..., v_{n}} \\
V_{TI_{input}} &= g(T, I_{input}) = {v_{1}, v_{2}, ..., v_{m}}
\end{split}
\end{equation}
where $f$ translates the text into visual concepts and $g$ translates the text and image into visual concepts, and $v_{i}$ are the individual visual concepts. For the sake of simplicity and clarity in the following discussion, we will refer to them as $V$. Humans, likewise, as soon as they receive a prompt to inspect are able to directly generate the visual concepts.

Once visual concepts are formulated, the human being immediately proceeds to examine whether these visual concepts are manifested in the image and how they interact with each other. This exploration is not a simple casual observation, but involves an unconscious questioning process in which the mind raises a multitude of implicit questions and then attempts to answer them using its inherent ability to understand visual content. The same idea drives the ViCE process.

In ViCE, the genesis of the process is marked by the generation of a group of "blind questions." These questions are derived from the use of previously formulated visual concepts. They are called blind because they are not based, unlike the refinement questions, on information that has been seen and processed from the generated image. This can be expressed as:
\begin{equation}
\label{eqn:question_generation}
Q_0 = q(V)
\end{equation}
Here, $Q_0$ denotes the initial set of blind questions, which are generated by applying function $q$ over the visual concepts, $V$.

Next, the image to be evaluated, $I$, is examined and, through reasoning, an effort is made to answer the questions and determine the presence of expected elements. This key step requires a comprehensive understanding and interpretation of the image.
\begin{equation}
\label{eqn:answer_generation}
A_0 = a(I, Q_0)
\end{equation}
In the equation above, $A_0$ are the initial answers. The function $a$ encapsulates the human-like capacity of the model to interpret and reason about the image to furnish responses to the blind questions. Hence, ViCE reflects the way a human mind functions when comprehending and evaluating visual content.

After obtaining the initial set of answers from the blind questions and having a clear understanding of the presence of the required visual concepts, the model (or human evaluator) has to make a decision $D$ if to request additional information to make some aspects clearer, or to close the process and make the final evaluation. If more information is needed, the model creates a new set of questions, known as "refinement questions". 

Such iterative process can be conducted indefinitely:
\begin{equation}
\label{eqn:refinement_loop}
\begin{split}
Q_i &= q'(V, Q_0, A_0, ..., Q_{i-1}, A_{i-1}) \\
A_i &= a(I, Q_i)
\end{split}
\end{equation}

Finally, the evaluation score is computed, using all the questions, answers, and initial visual concepts:

\begin{equation}
\label{eqn:evaluation_generation}
E = h(T, Q_0, A_0, Q_1, A_1, ..., Q_i, A_i)
\end{equation}

This recurrent process mirrors the human strategy for assessing a generated image, with each phase performing a crucial function in the overall evaluation.

\section{Implementation}

\begin{figure*}
    \centering
    \includegraphics[width=\textwidth]{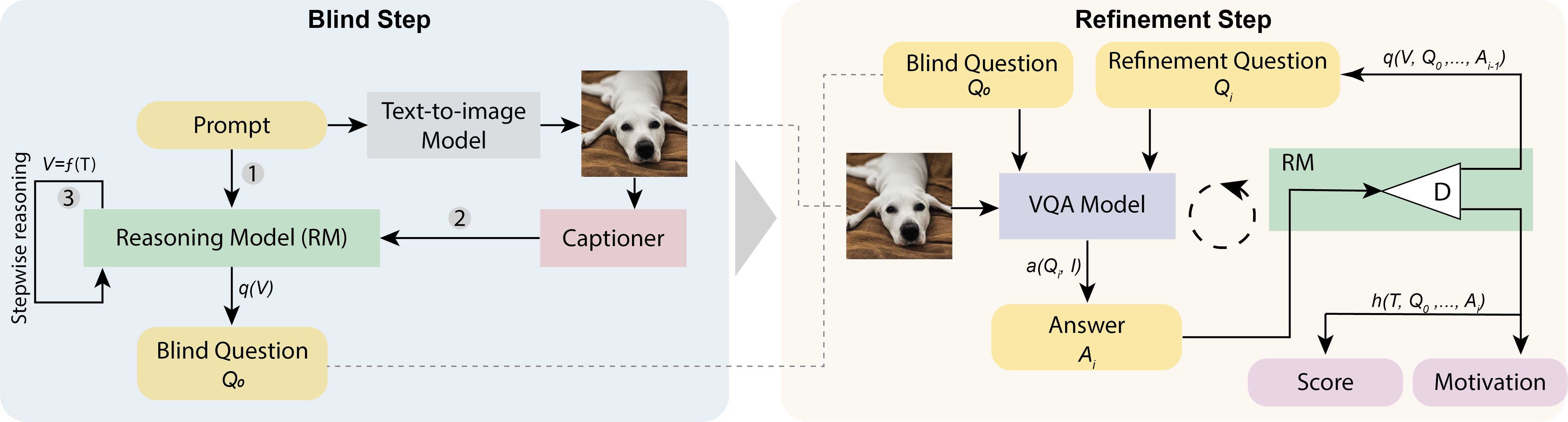}
    \caption{Visual Concept Evaluation Pipeline}
    \label{fig:full_pipeline}
\end{figure*}

In our implementation, we establish a pipeline that integrates various models, each having a specific function corresponding to the steps delineated in the previously mentioned equations \ref{eqn:vo_from_text}, \ref{eqn:question_generation}, \ref{eqn:answer_generation}, \ref{eqn:evaluation_generation}. The objective is to construct an autonomous system capable of evaluating synthetic generated images.

We have integrated a Large Language Model, specifically the GPT-3.5-turbo~\cite{openai_gpt-4_2023}, for the reasoning process. We refer to this agent as the "Reasoning Model". This choice was driven by our intent to simulate human-like reasoning, which is inherently stepwise, a characteristic LLMs readily adapt to~\cite{yao_tree_2023}. Stepwise reasoning consists in first asking the model what it expects to find in an image generated with that prompt and on what criteria it should evaluate the effectiveness of the generation. Only thereafter the actual questions are generated. As part of our future work, we plan to conduct a comparative study using different-sized LLMs and different stepwise reasoning approaches to verify and measure any impact on the results. 
To aid the model's reasoning process, we supplemented it with an image caption. Our analysis indicated that particularly for images that deviated from the expected generation, providing an image description significantly improved the model's capability to pose pertinent questions in the subsequent stages.

The question-generation phase unfolds in several steps. Initially, we set the model to generate a fixed number of questions (N=15 in our experiments) based on the image prompt and the expected visual concepts. 
Emulating the human evaluation process, the Reasoning Model may seek additional information to refine its understanding of the image. Therefore, as illustrated in Fig \ref{fig:full_pipeline}, the model is queried after the initial response phase about whether it requires further information. This triggers a refinement cycle featuring an iterative exchange of questions and answers until the model is satisfied with its comprehension of the image.

The questions span across semantic and qualitative aspects of the image, examining the presence or absence of objects described in the prompt, their interrelations, and qualitative characteristics. It is noteworthy how our Reasoning Model, in its pursuit to mirror human cognition, transcends the mere `words' in the prompt. It comprehends the necessity to validate whether the objects are in the correct semantic relationship. For instance, in response to the prompt `a vase of flowers', the model not only confirms the presence of the vase and the flowers but also verifies that the flowers are indeed in the vase, the vase is positioned on a surface, and the setting is congruent.

The responsibility of visual image analysis and generation of answers is vested in the Visual Question Answering (VQA) model. For our implementation, we utilized the BLIP2 model~\cite{li2023blip2}, built from the \texttt{salesforce-lavis} library~\cite{li2022lavis}. As with the LLM, we intend to investigate the influence of the VQA model on the final output in our future endeavors. The capabilities of the VQA model are crucial as they enable the Reasoning Model to construct a detailed image schema that informs the subsequent question cycles and, ultimately, the final evaluation.

\section{Experiments}

Our experimental setup focused on evaluating images that were generated from textual prompts. The key objective was to determine the extent of alignment between the evaluation scores procured from the Visual Concept Evaluation (ViCE) model and those rendered by human evaluators.

In the beginning, we used the Stable Diffusion 2 model to generate images, utilizing prompts extracted equally from a variety of datasets~\cite{feng2022training,cho2022dall,saharia2022photorealistic,lin2014microsoft} for a total number of 1000 images. The task given to the external evaluators was to assess the level of consistency between the prompt and the generated image by scoring on a scale from 0 to 10.

We compared the evaluation scores from our ViCE model with automated metrics such as CLIPScore~\cite{hessel2021clipscore} and BLIP-ITC/ITM~\cite{li2023blip2}, along with other model-based evaluation techniques like LLMScore. CLIPScore and BLIP-ITC measure the distance between the embedding of the generated image and the embedding of the prompt. BLIP ITM has an additional network submodule that outputs a probability of matching.
% It is important to underline that metrics like the Inception Score (IS) and Frechet Inception Distance (FID), which can't be directly applied to a comparison between the prompt and the generated image, were excluded from our methodology. These metrics, designed to compare two images, may be utilized in Image Targeted Editing (ITE) tasks. In forthcoming sections, we will explore further how our evaluation methodology can be extended to ITE tasks.

\subsection{Comparison with Human Evaluation}
We conducted a comparative study between the scores derived from human evaluations and the calculated metrics. This comparison was accomplished using two correlation coefficients: Spearman's rank correlation coefficient and Pearson's correlation coefficient; additionally, we also used the Bland-Altman plot to illustrate the agreement between human and model-derived scores. More in detail, we employed:

\begin{itemize}
    \item Spearman's Rank Correlation Coefficient: This non-parametric measure assesses the strength and direction of the relationship between two ranked variables. As it is less sensitive to outliers and does not assume a linear relationship, it is ideal for comparing ordinal variables.
    \item Pearson's Correlation Coefficient: This measure evaluates the linear correlation between two continuous variables.
    \item Bland-Altman Plot: This graphical method measures the agreement between two different ways of measuring a variable (in our case, human and model-derived evaluations). The plot showcases the difference between the two measurements against their average. 
\end{itemize}

\subsection{Results}

\begin{table}[t]
\centering
\begin{tabular}{|l|c|c|}
\hline
\textbf{Model} & \textbf{Pearson} & \textbf{Spearman} \\
\hline
CLIPscore                   & 0.19467                & 0.17452              \\
BLIP ITM                    & 0.19404                & 0.18752              \\
BLIP ITC                    & 0.26943                & 0.25421              \\
LLMScore                    & 0.29264                & \textbf{0.34065}              \\
\specialrule{.1em}{.05em}{.05em} % This creates a thicker line
ViCE\_5                      & 0.25221                & 0.24981              \\
ViCE\_blind                  & 0.27547                & 0.28325              \\
ViCE                         & \textbf{0.33249}                & 0.32762              \\
\hline
\end{tabular}

\vspace{0.3cm}
\caption{Comparison of Evaluation Models. All metrics report p-value lower than 0.05, indicating statistically significant correlations. ViCE\_5 applies the same pipeline with 5 questions and without refinements questions; ViCE\_blind only uses the blind questions, without refinement. }
\vspace{-1.2cm}
\label{tab:results}
\end{table}

The results presented in Table~\ref{tab:results} reflect the evaluation carried out across several datasets, thus providing an overall score that accounts for different domains across these datasets. Notably, both LLMScore and ViCE significantly surpass all other automated metrics. An interesting observation is that while LLMScore performs better in terms of Spearman correlation, ViCE excels in Pearson correlation.

This outcome warrants a brief exploration. Spearman correlation evaluates the monotonic relationship between the two datasets, while Pearson correlation assesses the linear relationship. Therefore, ViCE's superiority in Pearson correlation might suggest a better linear relationship with the human scores.

Moving forward, our goal is to further refine  ViCE by introducing an initial caption similar to the strategy employed by LLMScore. 
We envision that incorporating local and global descriptors, drawing from the methodology of GRIT~\cite{wu_grit_2022}, could improve the effectiveness of ViCE.

Additionally, in Table~\ref{tab:results}, we include the results from the ViCE model with only 5 initial questions (`ViCE\_5') and without the refinement questions (`ViCE\_blind'). Our hypothesis, which is supported by these results, suggests that reducing the number of questions or completely removing the refinement process prevents the model from effectively reason and use the visual feedback, two elements that even humans leverage during evaluation.

\begin{figure*}[t]
    \centering
    \begin{subfigure}{0.33\textwidth}
        \centering
        \includegraphics[width=\linewidth]{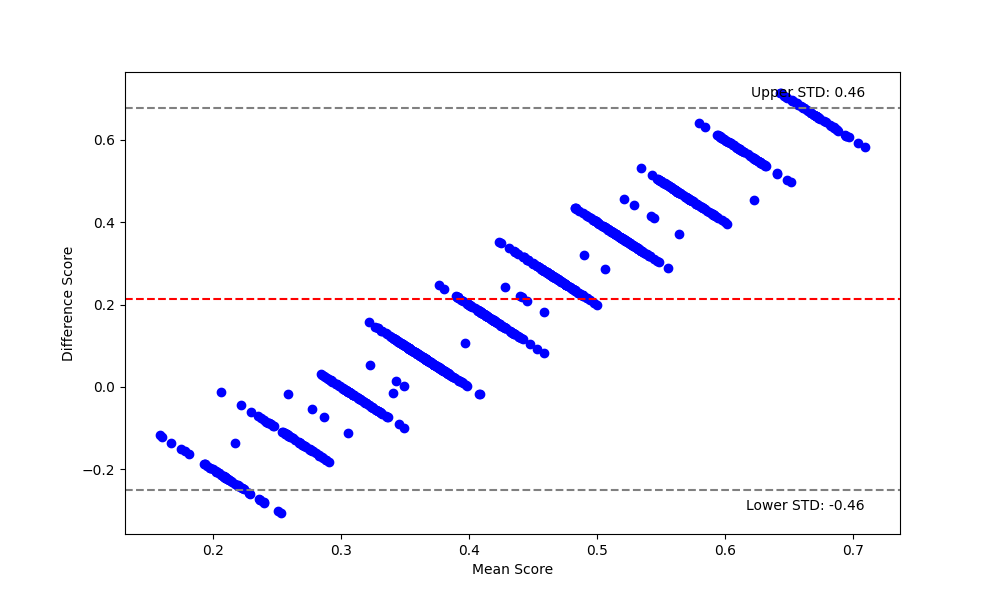}
        \caption{CLIPScore}
    \end{subfigure}
    \begin{subfigure}{0.33\textwidth}
        \centering
        \includegraphics[width=\linewidth]{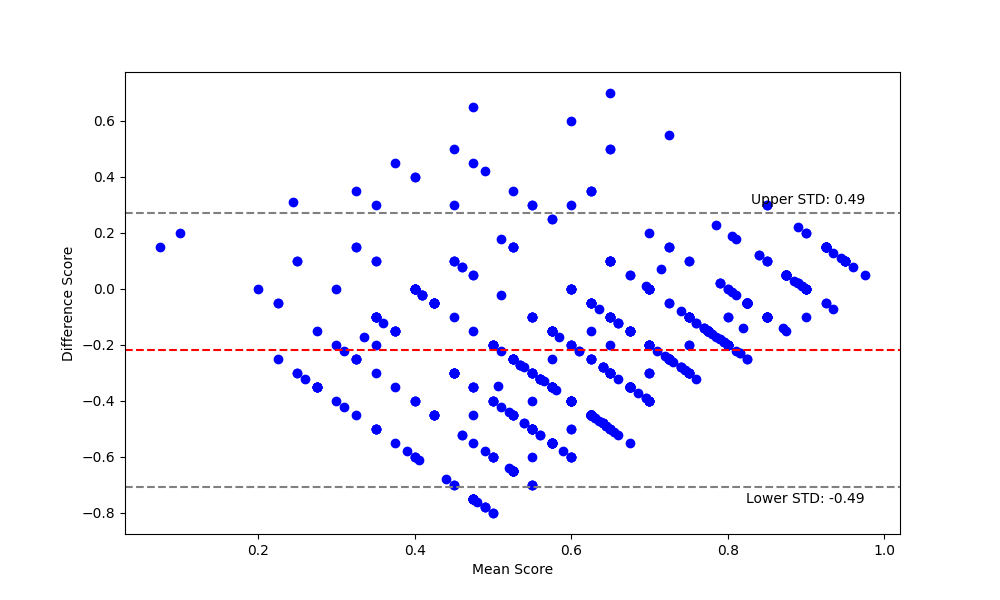}
        \caption{LLM\_score}
    \end{subfigure}
    \begin{subfigure}{0.33\textwidth}
        \centering
        \includegraphics[width=\linewidth]{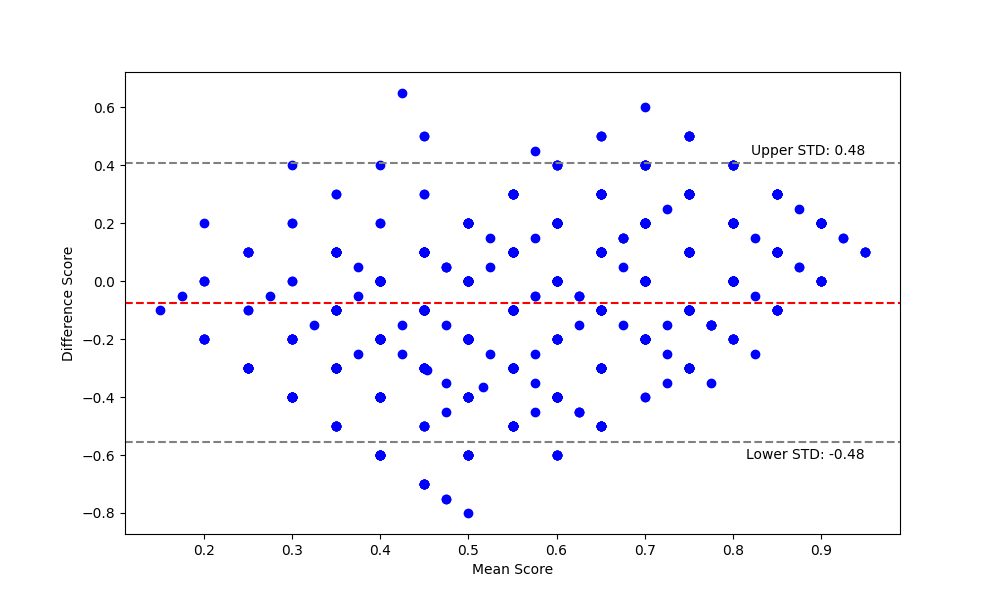}
        \caption{ViCE}
    \end{subfigure}
    \caption{Bland-Altman plots for different automated metrics.}
    \label{fig:bland_plots}
\end{figure*}

\begin{figure*}[t]
    \centering
    \includegraphics[width=\textwidth]{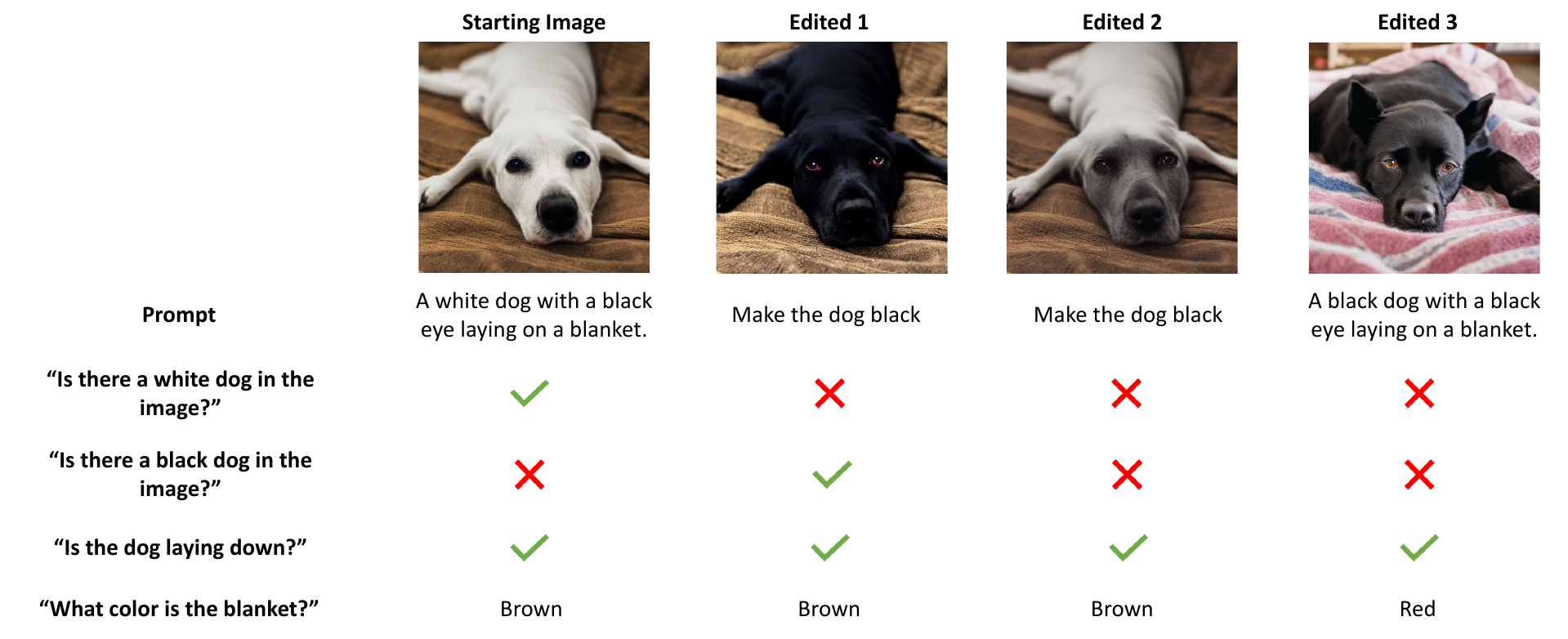}
    \caption{ViCE applied to the ITE task, whereby an LLM generates context-specific queries to assess the quality of the edit. The variation in the generated responses offers insights about the effectiveness of the edit operations.}
    \label{fig:ITE}
\end{figure*}

In Figure~\ref{fig:bland_plots} we report the Bland-Altman graphs~\cite{bland1986statistical}, an established method to visualize the differences between two measurement techniques. In our setup, it provides a visual representation of the agreement between a standard reference measure (i.e. the human evaluation) and an automated metric of interest (i.e. one amongst CLIPScore, LLM\_score, and our proposed ViCE), while simultaneously exposing any potential biases in the assessment.

It can be observed how CLIPScore's data fluctuates in a narrow range. Conversely, LLM\_score tends to assign higher scores than human assessments, a fact that indicates a potential overestimation of image quality. On the other hand, ViCE shows a balanced distribution, indicating a closer alignment with human evaluations and suggesting it can indeed offer a more reliable method for automatic evaluation.

\section{Extension to ITE}

Expanding on our prior discussions, we suggest that the Visual Concept Evaluation (ViCE) approach is not confined to image generation but can be extended to Image Targeted Editing (ITE). In the ITE task, the input comprises both an image and a descriptive prompt, with the latter containing instructions for the desired semantic changes to be applied to the image. Such modifications, rather than being stylistic adjustments, involve content alterations that touch upon only a section of the original image.

In the (recent) past, these models required an explicit mask to pinpoint the image section to be modified~\cite{meng_sdedit_2022, avrahami_blended_2022}. 
However, the currently available large Vision-Language models are now capable of autonomously identifying the region for modification~\cite{kawar2023imagic, zhang_hive_2023, brooks2023instructpix2pix}. 

Still, the evaluation of such a task requires human evaluators to identify which parts of the image should remain untouched and which should be altered, and then to evaluate the precision of the implemented changes.

In this context, visual concepts can be divided into three distinct sets:
\begin{enumerate}
    \item $V_{\text{remain}}$: Visual concepts that should be kept from the original image;
    \item $V_{\text{remove}}$: Visual concepts that should no longer be present;
    \item $V_{\text{add}}$: Visual concepts that should be added to the output image.
\end{enumerate}

Thus, the set of visual concepts $V$ to be checked for in the edited image compounds to:
% 1. $V_{\text{remain}}$: Visual concepts that will remain within the image.
% 2. $V_{\text{remove}}$: Visual concepts that will no longer be present.
% 3. $V_{\text{add}}$: Visual concepts that will be added.
\begin{equation}
V = V_{\text{remain}} - V_{\text{remove}} + V_{\text{add}}
\end{equation}
Through the reasoning process, questions related to the visual concepts that will be modified can be formulated, and the responses can subsequently be used to evaluate the effectiveness of the modification. 

Visual concepts belonging to $V_{\text{remain}}$ are expected to stay constant, and any change in the responses associated with these concepts would suggest that the portion of the image meant to be preserved has been altered. An illustrative example of this scenario can be found in Figure~\ref{fig:ITE}.

\section{Conclusions}
This work marks a initial step towards mirroring human reasoning when it comes to
synthetic image evaluation.
% multimedia data evaluation. 
We have devised an approach that acknowledges both explicit and implicit facets of human cognition, creating a close alignment with human judgment. 

This bold venture aims to narrow the cognitive gap between AI and humans, thereby advancing towards a more nuanced and reliable image evaluation methodology.

\section{Acknowledgments}
This work was supported by the MUR PNRR project FAIR - Future AI Research (PE00000013) funded by the NextGenerationEU, the PRIN project CREATIVE (Prot. 2020ZSL9F9), and the Horizon Europe project ``European Lighthouse on Safe and Secure AI (ELSA)'' (HORIZON-CL4-2021-HUMAN-01-03), co-funded by the European Union (GA 101070617). The work of JS has been partially funded by \href{http://www.ipazia.com}{Ipazia S.p.A.}.

% \section{Appendices}

\bibliographystyle{ACM-Reference-Format}
\bibliography{bibliography}

%\appendix

%Appendix content

\end{document}